\documentclass[11pt,twocolumn]{article}
\usepackage[utf8]{inputenc}

\usepackage[margin=1in]{geometry}
\usepackage{setspace}
\usepackage{amsmath,amssymb}
\usepackage{newunicodechar}
\usepackage[inline]{enumitem}

\usepackage{tabularx}
\usepackage{makecell}
\usepackage[table]{xcolor}
\usepackage{booktabs}
\usepackage{listings}
\usepackage{graphicx}
\usepackage{tikz}
\usetikzlibrary{arrows.meta,positioning,shapes.geometric}

\usepackage[numbers]{natbib}   
\usepackage{hyperref}
\usepackage{footmisc}

\usepackage{enumitem}

\title{Small Edits, Big Consequences: Telling Good from Bad Robustness in Large Language Models}
\author{Altynbek~Ismailov (altyni@gmail.com) \and Salia~Asanova (saliia.asanova@berkeley.edu)}

\begin{document}
\maketitle


\begin{abstract}
LLMs are now deployed in settings where overlooking a single word can have safety or financial consequences. Yet we also expect them to forgive our stray typos or slip-ups. To understand the boundaries between “good” and “bad” sensitivity, we test frontier models with tiny prompt edits to see when they stubbornly reuse a canned solution and when they truly adapt. 

We compile 50 LeetCode problems and apply three cumulative perturbation regimes that differ in how much they \textbf{should} matter:

\begin{enumerate*}[label=(\roman*)]
    \item \textbf{Progressive underspecification} — deleting 10 \% of tokens per step (mostly benign noise to be ignored or clarified);
    \item \textbf{Lexical flip} — swapping a pivotal quantifier  (\emph{max}$\!\to\!$\emph{min}) that \emph{must} be honoured;
    \item \textbf{Jargon inflation} — replacing a common noun with an obscure technical term, whose importance is ambiguous.
\end{enumerate*}

Six frontier models—including three ``reasoning-optimised’’ variants—solve each mutated prompt; their Python code is executed against the \emph{original} test suites to reveal whether they regressed to the baseline solution or adapted.  
Across 11 853 generations we observe a double asymmetry:

\begin{itemize*}
  \item \textbf{Robust to noise when they should fail.}  
        Even after losing 90 \% of the prompt, models still pass in \textbf{85 \%} of cases, signalling over-robustness to underspecification.
  \item \textbf{Insensitive to meaning when they should adapt.}  
        A single quantifier flip changes the task semantics, yet only \textbf{54 \%} of generations react; ``reasoning’’ variants are \emph{less} sensitive than their baselines.
\end{itemize*}

Jargon inflation sits in the middle yielding a 56 \% pass-through rate that exposes the models’ difficulty in judging contextual importance.  

Contemporary LLMs blur the line between noise and meaning-changing edits: frequently they treat both as ignorable. Masking salient anchors, such as descriptive function name, forces re-evaluation. We therefore advocate evaluation and training protocols that measure and reward \emph{differential sensitivity}: models should remain steady under benign noise yet adapt (or refuse) when an edit flips task semantics.

\end{abstract}

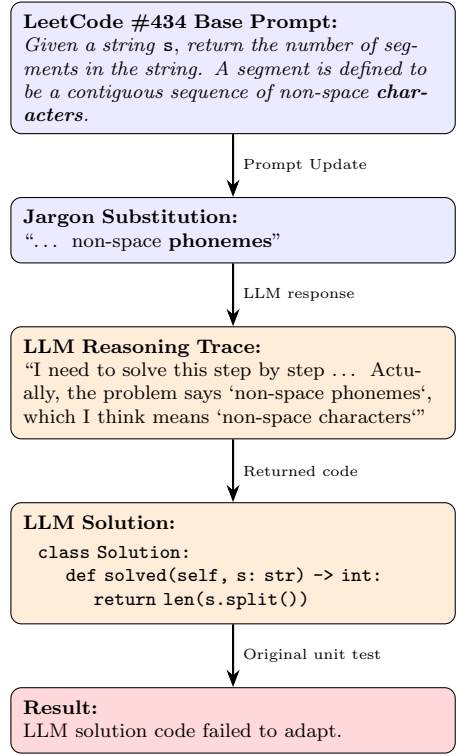
\begin{figure}[t!]
\centering
\begin{tikzpicture}[
  scale=.93, transform shape,
  prompt/.style={draw,fill=blue!8,rounded corners,
                 text width=6cm,align=left,
                 font=\scriptsize,inner sep=5pt},
  resp/.style={draw,fill=orange!15,rounded corners,
               text width=6cm,align=left,
               font=\scriptsize,inner sep=5pt},
  test/.style={draw,fill=red!15,rounded corners,        
               text width=6cm,align=left,
               font=\scriptsize,inner sep=5pt},
  arr/.style={-Stealth,thick},
  node distance=.9cm
]
\node[prompt] (base) {\textbf{LeetCode \#434 Base Prompt:}\\
\emph{Given a string} \verb|s|, \emph{return the number of segments in the string. A segment is defined to be a contiguous sequence of non-space \textbf{characters}.}};
\node[prompt,below=of base] (jargon) {\textbf{Jargon Substitution:}\\
``… non-space \textbf{phonemes}''};

\node[resp,below=of jargon] (reason) {\textbf{LLM Reasoning Trace:}\\
``I need to solve this step by step … Actually, the problem says `non-space phonemes`, which I think means `non-space characters`''};

\node[resp,below=of reason] (code) {\textbf{LLM Solution:}\\[.5em]
\begin{tabular}{l}
\verb|class Solution:|\\
\verb|    def solved(self, s: str) -> int:|\\
\verb|        return len(s.split())|
\end{tabular}};

\node[test,below=of code] (test) {\textbf{Result:}\\
LLM solution code failed to adapt.};
\draw[arr] (base)   -- node[right,font=\tiny]{Prompt Update} (jargon);
\draw[arr] (jargon) -- node[right,font=\tiny]{LLM response} (reason);
\draw[arr] (reason) -- node[right,font=\tiny]{Returned code}   (code);
\draw[arr] (code)   -- node[right,font=\tiny]{Original unit test}    (test);
\end{tikzpicture}
\caption{Jargon-inflation stress test on LeetCode \#434. Replacing “characters” with a specialised term (e.g., “phonemes”) still yields the canonical whitespace-split solution, revealing full regression to the original task even under new linguistic level constraint.}
\label{fig:fig1}
\end{figure}

\section{Introduction}\label{sec:intro}
\noindent\textit{“The devil is in the details.”}  
Few domains illustrate this maxim more sharply than large-language-model
(LLM) deployment.  Modern systems can draft contracts, file bug-fixes,
even suggest medical plans—yet one missing negation or swapped
quantifier can reverse the intended action.  At the same time, we
\emph{want} models to overlook typos, stray commas, or irrelevant
parentheticals.  In other words, \textbf{robustness should be selective}:
ignore benign noise, but attend to meaning-critical edits.

\vspace{1mm}
\paragraph{Research question.}
\emph{Where do today’s LLMs draw the line between noise that can be
safely ignored and edits that must change the answer?}

\vspace{1mm}
\paragraph{Good vs.\ bad sensitivity.}
We operationalise the distinction with three perturbation families:

\begin{enumerate*}[label=(\roman*),itemjoin*={;\ },afterlabel=\hspace{0.3em}]
    \item \textbf{Progressive underspecification}: delete 10\,\% of the
          prompt tokens at each step—harmless noise that the model should
          ignore or ask clarification about;
    \item \textbf{Lexical flip}: swap a pivotal quantifier
          (e.g.\ \emph{max}$\!\to$\emph{min})—a change the model
          \emph{must} respect;
    \item \textbf{Jargon inflation}: replace a commonplace noun
          (“characters”) with an obscure term (“graphemes”,
          “phonemes”)—importance is ambiguous, a grey zone.
\end{enumerate*}

\vspace{1mm}
\paragraph{Method at a glance.}
Six leading models—\textbf{GPT-4.1}, \textbf{o4-mini-high},
\textbf{Sonnet-4}, \textbf{Sonnet-4-thinking}, \textbf{Gemini-2.5-Flash},
\textbf{Gemini-2.5-Pro}—solve 50 LeetCode problems, each mutated across
ten steps in the three families.  
Generated Python is executed \emph{against the original test suite} to
see whether the model \emph{regresses} to the canonical solution
(pass) or \emph{diverges} (fail).  
Masking the function name lets us measure reliance on that single anchor.

\vspace{1mm}
\paragraph{Main empirical patterns.}
Across the full set of 11\,853 model–prompt pairs three effects appear
consistently:

\begin{enumerate}[label=\textbf{\arabic*.},leftmargin=1.5em,nosep]
    \item \textbf{Identifier anchoring.}  
          A single descriptive function name (and, to a lesser extent,
          the title and signature) acts as a compass.  
          With the name intact, models survive extreme
          underspecification; masking it with a neutral token
          (\verb|solved|) cuts pass-rates by \textbf{15–21\,pp} across
          \textbf{Progressive Deletion} (PD), \textbf{Jargon Inflation} (JI),
          and \textbf{Lexical Flip} (LF).  
          In other words, removing the anchor \emph{reduces} “false
          robustness’’ to missing detail but also makes the model more
          willing to adapt when semantics truly change.

    \item \textbf{Selective robustness gone awry.}  
          Anchored models are correctly tolerant of massive noise
          (PD-Unmasked: \textbf{85 \%} pass on average) \emph{but} they also ignore meaning-critical edits
          (LF-Unmasked: \textbf{54 \%} adapt; JI-Unmasked:
          \textbf{44 \%}).  
          The mechanism that filters benign perturbations therefore
          masks dangerous ones.

    \item \textbf{Divergent internal heuristics.}  
          Once anchor cues disappear, the six models disagree on
          pass/fail for the same mutated prompt in \textbf{30–40 \%} of
          cases, revealing heterogeneous task assumptions beneath the
          uniform API surface.
\end{enumerate}

The picture that emerges: today’s LLMs privilege high-probability,
well-rehearsed templates unless a constraint is \emph{both} concise and
explicitly foregrounded.

\vspace{1mm}
\paragraph{Roadmap.}
Section~\ref{sec:data} describes the dataset;  
Section~\ref{sec:exp} details the perturbation protocol;  
Section~\ref{sec:results} presents findings on good vs.\ bad
sensitivity;  
Section~\ref{sec:limits} discusses limitations; and  
Section~\ref{sec:related} situates our work in the broader literature.
\section{Dataset}\label{sec:data}

\paragraph{Source corpus.}

We build on the publicly released \textsc{LeetCodeDataset} of \citet{Xia2025}, a collection of {2\,869} programming problems drawn from LeetCode, each paired with the platform’s reference test suite.  The dataset offers a uniform schema (title, statement, starter code, tests, problem description) that is well-suited to prompt-construction studies.

\paragraph{Pre-processing.}
Before applying our perturbations we perform three cleaning steps:

\begin{enumerate}[label=(\alph*),leftmargin=1.4em,nosep]
    \item \textbf{Strip usage hints.}  We delete the \textit{Examples} and \textit{Constraints} blocks so that later token deletions or word swaps cannot contradict hard-coded I/O pairs.
    \item \textbf{Normalise whitespace.}  Redundant line-breaks and Markdown artefacts are collapsed to a single space; code snippets are preserved verbatim.
    \item \textbf{Canonicalise identifiers.}  Function names are processed so that they can be masked or replaced without touching surrounding text.
\end{enumerate}

\paragraph{Task subset.}
From the cleaned corpus we draw a sample of \textbf{50} problems.  A fixed random seed (\verb|seed=42|) guarantees reproducibility.

\paragraph{Query Prompt.}
We used the same query prompt as in the dataset with our own track specific updates to the problem description (PD, JI, LF) and starter code (masked/unmasked). A query prompt asks LLMs to solve the LeetCode problem and return the response in a code format with additional comments.(Appendix A)

\paragraph{Query properties.}
All query generations use temperature~$=0.7$ and the "best-of-1" decoding to minimise sampling noise

\section{Experimental Design}\label{sec:exp}

\noindent
Our study combines \textbf{three perturbation families} with
\textbf{two identifier modes} (unmasked vs.\ masked),
yielding six experimental tracks.  
Every track is applied to the 50-problem subset described in
Section~\ref{sec:data}, with up to \textit{ten} cumulative
perturbations per problem.

\paragraph{Perturbation budget.}
Each problem receive up to \textbf{10} successive edits.  
The full Cartesian product therefore comprises
\begin{equation}
\begin{aligned}
  & 50 \text{ problems} \times 10 \text{ steps}\\
  & \times 3 \text{ families} \times 2 \text{ identifier modes}\\
  & = 3\,000 \text{ distinct prompts per model,}
\end{aligned}
\end{equation}
or \textbf{11\,853} model–prompt pairs once early stopping for invalid
outputs is taken into account. Early stoppings were triggered when a duplicate problem description was generated (e.g. Successive Lexical flips undo each other).    (Table~\ref{tab:obs-counts}).

\subsection{Perturbation families}\label{sec:perturb}

\begin{enumerate}[label=\textbf{F\arabic*},leftmargin=1.6em,nosep]
    \item \textbf{Progressive deletion} :  
          delete \(10\,\%\) of the remaining tokens at each step, preserving
          punctuation and code.
    \item \textbf{Jargon inflation}:  
          a meta-prompt substitutes \emph{one} commonplace term with an
          obscure technical synonym, tasked to be cumulative.
    \item \textbf{Lexical flip}:  
          swap a single quantifier, polarity, or other word
          (\emph{max}$\leftrightarrow$\emph{min}, etc.),
          again in cascading fashion.
\end{enumerate}

For Jargon inflation and Lexical flip tracks we used Claude-Sonnet-4 model for word substitutions and insertions using the prompts provided in Appendix B,C.   
\subsection{Identifier modes}\label{sec:mask}

Descriptive function names act as strong anchors, so we evaluate every
perturbation track in two variants:

\begin{itemize}[leftmargin=1.4em,nosep]
    \item \textbf{Unmasked} – original identifier retained.
    \item \textbf{Masked} – identifier replaced with the neutral
          placeholder \verb|solved| (all internal references updated).
\end{itemize}

\begin{table}[t]
\centering
\small
\setlength{\tabcolsep}{3pt}  
\caption{Experimental tracks: perturbation family $\times$ identifier mode.}
\label{tab:tracks}
\begin{tabular}{@{}lcc@{}}
\toprule
\textbf{Track} & \textbf{Family (Sec.~\ref{sec:perturb})}
               & \textbf{Identifier mode (Sec.~\ref{sec:mask})} \\
\midrule
PD--U & Progressive deletion & Unmasked \\
PD--M & Progressive deletion & Masked   \\
JI--U & Jargon inflation     & Unmasked \\
JI--M & Jargon inflation     & Masked   \\
LF--U & Lexical flip         & Unmasked \\
LF--M & Lexical flip         & Masked   \\
\bottomrule
\end{tabular}
\end{table}

\subsection{Model roster}\label{sec:models}

We test six frontier LLMs, grouped by whether the vendor advertises an
explicit chain-of-thought decoding mode:

\begin{description}[labelsep=0.6em,leftmargin=1.6em,style=nextline]
  \item[\textbf{Baseline variants:}]
        \hfill\parbox[t]{\linewidth}{%
        \textbf{GPT-4.1}\,(OpenAI), \textbf{Claude-Sonnet-4}\,(Anthropic),
        \textbf{Gemini-2.5-Flash}\,(Google).}
  \item[\textbf{Reasoning-optimised variants:}]
        \hfill\parbox[t]{\linewidth}{%
        \textbf{o4-mini-high}\,(OpenAI), \textbf{Claude-Sonnet-4-thinking}\,(Anthropic),
        \textbf{Gemini-2.5-Pro}\,(Google).}
\end{description}

Note: We originally planned to use \textbf{o3}\,(OpenAI) model, but the endpoint was not readily available on OpenRouter, thus we opted to \textbf{o4-mini-high}\ model. The \textbf{o4-mini-high}\ model outperforms \textbf{o3}\ on CodeForces benchmark.   
\begin{table*}[t]
\centering
\small
\setlength{\tabcolsep}{4pt}
\caption{Observation counts by model and experiment\textsuperscript{*}}
\label{tab:obs-counts}
\begin{tabularx}{\textwidth}{@{}l*{6}{>{\centering\arraybackslash}X}r@{}}
\toprule
\textbf{Experiment} &
\makecell{Sonnet-4} &
\makecell{Sonnet-4\\thinking} &
\makecell{Gemini-2.5\\Flash} &
\makecell{Gemini-2.5\\Pro} &
\makecell{GPT-4.1} &
\makecell{o4-mini\\high} &
\textbf{Total} \\
\midrule
Progressive deletion & 1,098 &   0 & 1,098 &   0 & 1,098 &   0 & 3,294 \\
Jargon inflation     & 1,044 & 522 & 1,044 & 522 & 1,044 & 522 & 4,698 \\
Lexical flip         &   858 & 429 &   858 & 429 &   858 & 429 & 3,861 \\
\midrule
\textbf{Total}       & \textbf{3,000} & \textbf{951} & \textbf{3,000} &
                       \textbf{951} & \textbf{3,000} & \textbf{951} &
                       \textbf{11,853} \\
\bottomrule
\end{tabularx}

\vspace{2pt}
\footnotesize\textsuperscript{*}\,Because of cost, we ran the reasoning variants only on the scenario that
yields the \emph{lowest} pass-rates, giving conservative estimates of
any accuracy gap.
\end{table*}

\section{Experimental Results}\label{sec:results}

\begin{figure*}[t]
    \centering
    \includegraphics[width=.9\textwidth]{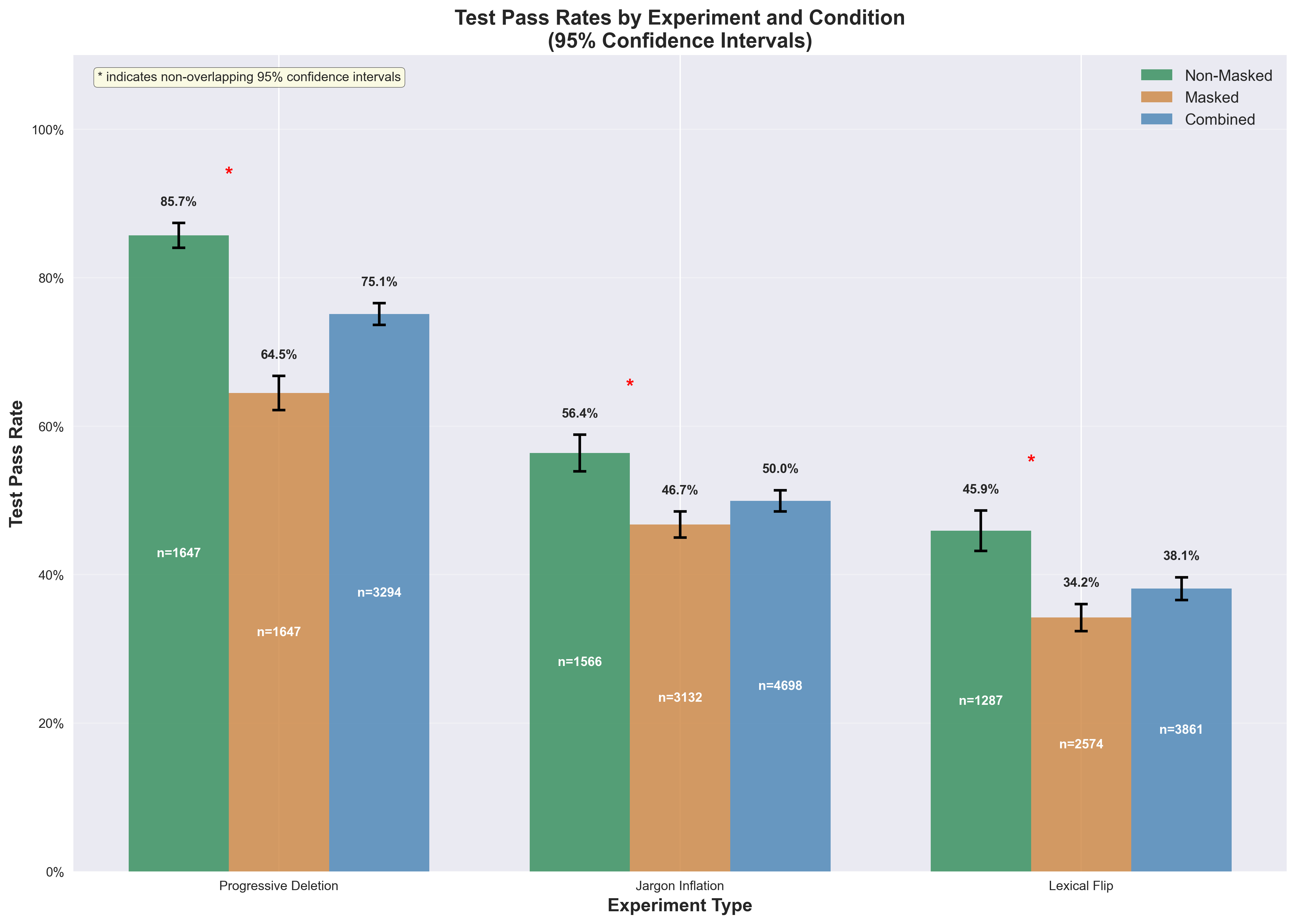}
    \caption{Pass-rates by perturbation family and identifier mode.
             Whiskers denote 95\,\%~confidence intervals.}
    \label{fig:pass-rate}
\end{figure*}

Figure~\ref{fig:pass-rate} reports pass-rates for the six models on the
six evaluation tracks (\S\ref{sec:exp}).  Four observations frame the
rest of the analysis.

\vspace{1mm}
\paragraph{(i) Extreme underspecification looks “safe’’—until the anchor goes.}
With the descriptive function name intact (\textbf{PD–Unmasked}) every
model keeps a median pass-rate above \textbf{85\,\%} after deleting
90 \% of the prompt.  Once the name is masked (\textbf{PD–Masked}),
the median drops to \textbf{65\,\%}.  In short, residual cues are
sufficient for template reuse only while the anchor is present.

\vspace{1mm}
\paragraph{(ii) Identifiers outweigh everything else.}
Across the three perturbation families, masking a single name reduces
pass-rate by \textbf{10–21\,pp} ($p<.001$).  The effect is larger than
that of deleting tens of tokens or injecting advanced jargon,
highlighting the outsized role of identifier anchoring.

\vspace{1mm}
\paragraph{(iii) Concise semantic pivots break templates; jargon does not.}
Swapping one polarity word (\textbf{LF}) cuts median pass-rate to
\textbf{46\,\%} in the unmasked setting and to \textbf{34\,\%} when the
identifier is already masked.  By contrast, \textbf{JI} still passes
in \textbf{56\,\%} and \textbf{46\,\%} of cases respectively: models tend to ignore obscure terms that
drift too far from familiar structural patterns, whereas a single
quantifier flip changes the algorithmic logic they cannot safely
overlook.

\subsection{Do ``reasoning'' variants help?}\label{sec:thinking}

\begin{figure*}[t]
    \centering
    \includegraphics[width=.75\textwidth]
        {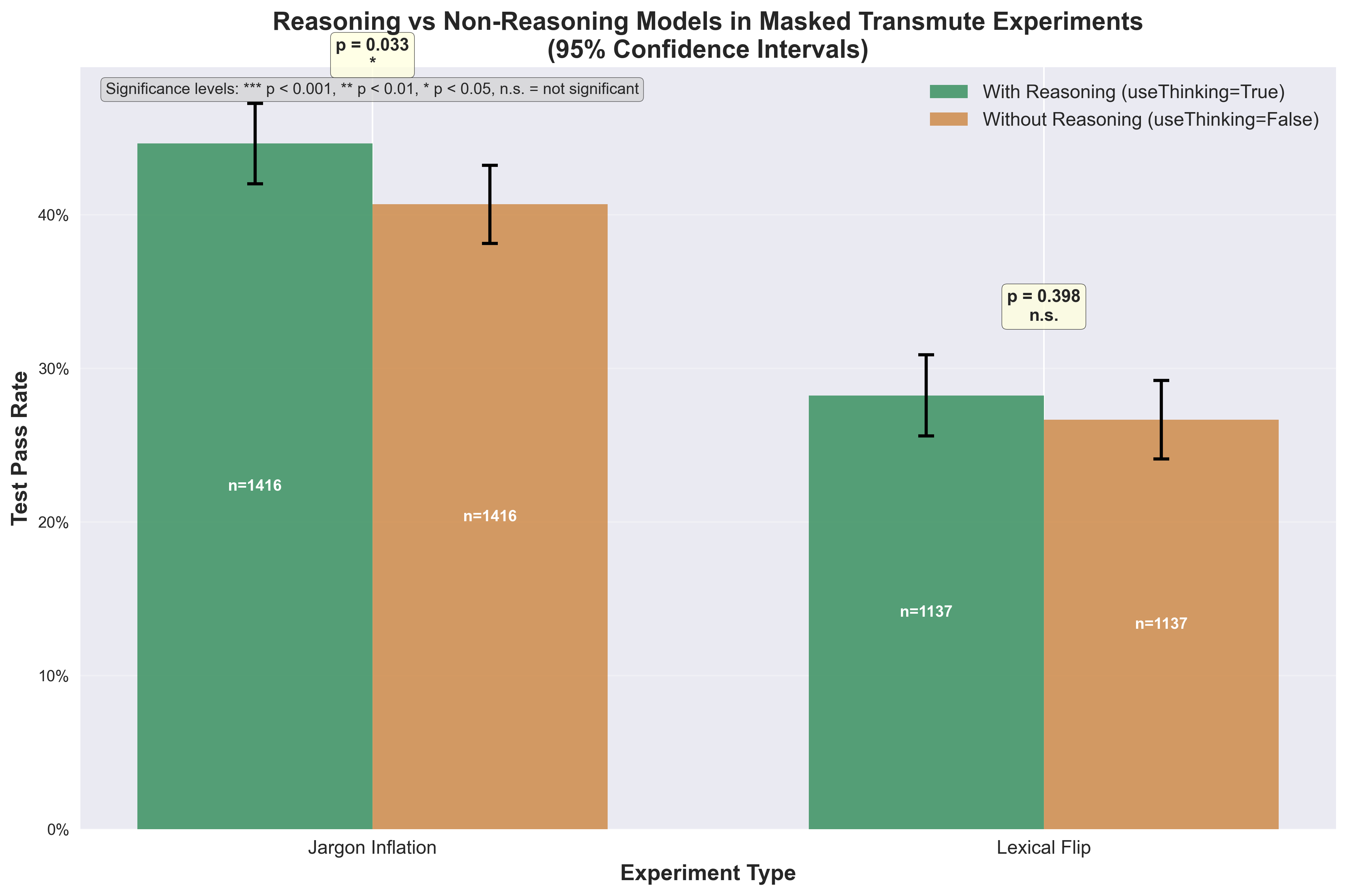}
    \caption{Pass-rate difference (\%) between reasoning-optimised and
             baseline variants for each track.  
             Error bars show 95\,\% confidence intervals.}
    \label{fig:thinking-gap}
\end{figure*}

Vendor-flagged “reasoning’’ endpoints—\textbf{o4-mini-high},
\textbf{Claude-Sonnet-4-thinking}, and \textbf{Gemini-2.5-Pro}—track their
silent counterparts almost point-for-point
(Figure~\ref{fig:thinking-gap}).  
Because of cost, we ran the reasoning variants only on the scenario that
yields the \emph{lowest} pass-rates, giving conservative estimates of
any accuracy gap.

\begin{itemize}[leftmargin=1.5em,nosep]
  \item \textbf{Jargon-Inflation (JI).}  
        Pass-rates for the reasoning variants are \emph{higher},
        indicating \textbf{+4.0\,pp} more regressions to the canonical
        answer (\emph{p}\,$<0.05$).
  \item \textbf{Lexical-Flip (LF).}  
        The average difference is a non-significant \(+1.6\,\)pp.
\end{itemize}

Thus, exposing a chain-of-thought does \emph{not} improve—and may even
diminish—sensitivity to subtle prompt drift.

Taken together, the experiments reveal two broad tendencies in today’s
LLMs:

\begin{itemize}[leftmargin=1.5em,nosep]
  \item \textbf{They update their code more readily when the change is
        \emph{straightforward}.}  
        Swapping a single polarity word (\texttt{max}\,$\rightarrow$\,\emph{min})
        or inserting a crisp numeric bound leaves no ambiguity; most
        models abandon the old algorithm and implement the new
        requirement.

  \item \textbf{They also adapt when their usual anchors disappear.}  
        Stripping out scaffolding—descriptive function names,
        boiler-plate signatures, tell-tale tokens—removes the shortcut
        to a memorised template and forces the model to reread the
        prompt.  Pass-rates on the \texttt{masked} tracks rise precisely
        because the model can no longer cling to the canonical pattern.
\end{itemize}

Outside these two scenarios, default behaviour tends to reproduce the
highest-probability, well-rehearsed solution, leaving subtle or
ambiguous specification changes effectively invisible even to the
strongest reasoning LLMs.

\subsection{Case study: Jargon inflation on LeetCode \#434}\label{sec:case434}

\paragraph{Base task.}
\begin{quote}
\small
\textbf{Prompt.} \emph{Given a string} \verb|s|, \emph{return the number
of segments in the string.  A segment is defined to be a contiguous
sequence of non-space \textbf{characters}.}
\end{quote}

\paragraph{Jargon substitutions.}
Starting from the base prompt, the JI routine successively replaced the
noun “characters’’ with five increasingly specialised terms:

\begin{enumerate}[label=\arabic*.,leftmargin=1.4em,nosep]
  \item \emph{non-space \textbf{scalars}}
  \item \emph{non-space \textbf{glyphs}}
  \item \emph{non-space \textbf{graphemes}}
  \item \emph{non-space \textbf{phonemes}}
  \item \emph{non-space \textbf{morphemes}}
\end{enumerate}

The semantic scope widens from a Unicode code point (\emph{scalar}) to
linguistic units (\emph{phoneme}, \emph{morpheme}), implying that a
correct solution might need to parse bytes, font renderings, or even
speech symbols.

\paragraph{Outcome.}
Across all six models—baselines and “reasoning’’ variants alike—the
generated Python program remained \emph{byte-level}, splitting on
whitespace and counting substrings exactly as in the canonical solution.
Every variant passed the \emph{original} LeetCode test suite, signalling
full regression to the known task.

\paragraph{Illustrative rationale (reasoning model):}
\begin{quote}
\textit{“Let me understand this problem: 1. We have a string `s` 2. We need to count the number of segments in the string 3. A segment is defined as a contiguous sequence of non-space characters (morphemes)...”}
\end{quote}

\paragraph{Take-away.}
Even when confronted with terminology that \emph{should} invalidate the
byte-oriented template, current LLMs  
\textbf{(i)} map unfamiliar jargon back onto the nearest high-probability
interpretation, and  
\textbf{(ii)} regenerate the canonical algorithm verbatim—even in masked
settings.

\subsection{Step-by-step dynamics of prompt drift}\label{sec:stepwise}

\begin{figure*}[t]
    \centering
    \includegraphics[width=.9\textwidth]{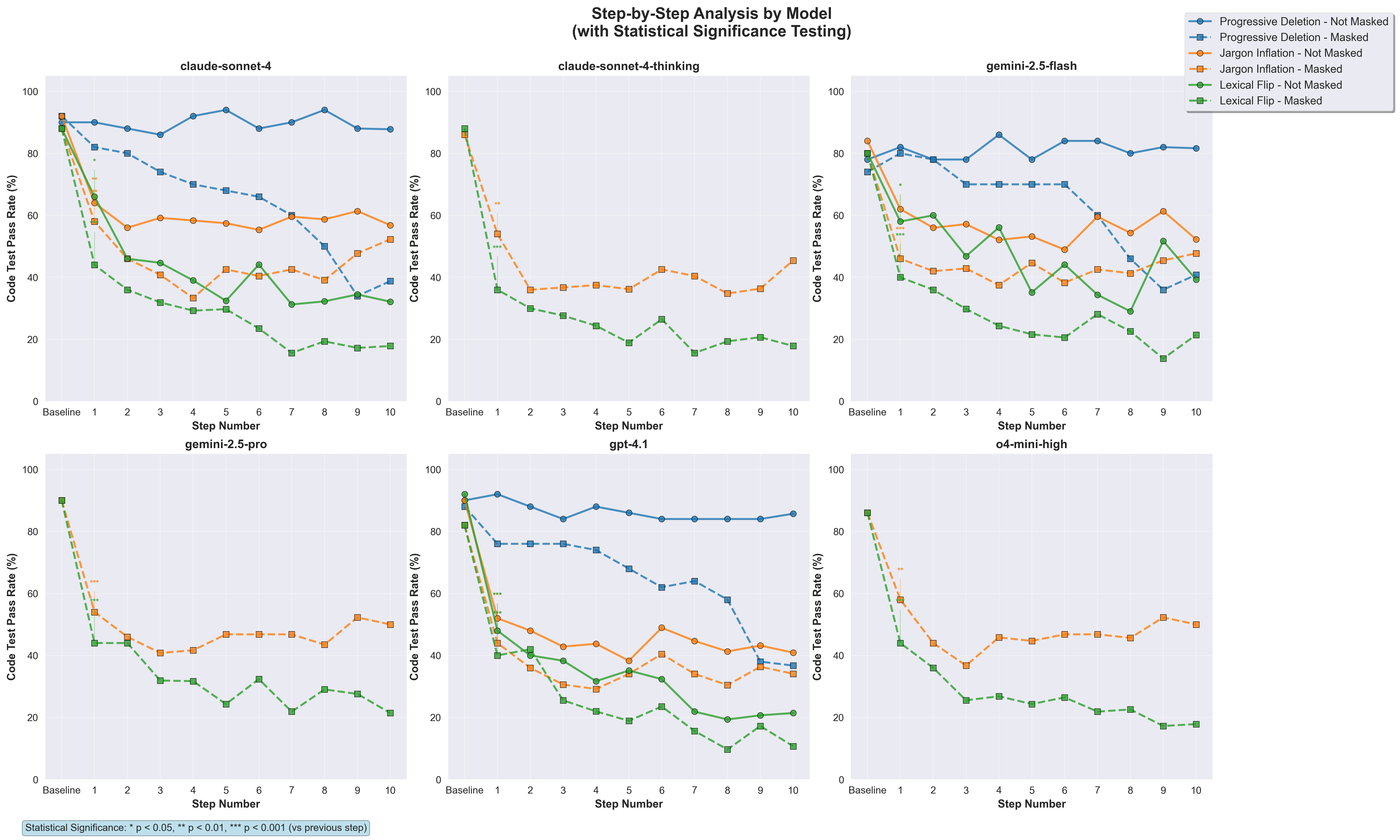}
    \caption{Per-model pass-rate trajectories.  Stars mark statistical
             significance vs.\ the previous step.}
    \label{fig:step-model}
\end{figure*}

Figures~\ref{fig:step-model} (model–specific panels) plot pass-rates from the untouched
\textsc{Baseline} through ten cumulative perturbation steps.  
Because Baseline itself is \textit{not} \(100\%\)  — around \(8-15\%\)  of problems remain
unsolved on average—each track starts with head-room to improve or degrade.  
Four detailed patterns emerge.

\paragraph{Progressive Deletion (PD).}
\textbf{Unmasked.}  Deleting chunks of text has almost no cost; in Claude- Sonnet - 4 and 
Gemini-2.5-Flash accuracy even ticks \emph{up} by 2–4 pp
after Steps 3-4.  All non-reasoning models keep \(\ge80\%\) pass-rate
throughout, at approximately their baseline rates, confirming that residual cues (function titles, identifiers, signatures) anchor the canonical solution.

\textbf{Masked.}  Removing the identifier reveals a gradual glide rather
than an immediate drop: the first seven steps shave only
1-4 pp per deletion and differences at each step are not statistically significant from previous one.
A dramatic collapse occurs only after \(\ge80\%\) of words vanish,
indicating that models remain surprisingly resilient—even without the
anchor.

\textit{Take-away:} PD shows genuine robustness to severe
underspecification; most failures arise only when both problem specification and anchor have essentially disappeared.

\paragraph{Jargon Inflation (JI).}
Both identifier modes suffer a sharp 25–35 pp hit at Step 1, yet the
curves flatten quickly afterwards.  
Masked and unmasked variants \emph{diverge} at first steps
(anchor makes a large difference early) but \emph{converge} by
Steps 6–8, when pass rate stabilises around \(50\%\).  
Technical obscurity thus matters most at first contact; as jargon
accumulates, both variants retreat to the same safe template.  
Reasoning models trace their silent counterparts with higher pass rates. 

\textit{Take-away:} Anchors modulate the immediate shock of jargon, but
longer sequences of obscure terms push every model back to the canonical
answer.

\paragraph{Lexical Flip (LF).}
Here masked and unmasked trajectories run almost in parallel, separated
by only 3–6 pp, yet both fall far faster than in PD or JI.  
Each cumulative polarity swap removes another rung of the original
algorithm; by Step 10 pass-rates drop below \(30\%\) for most baselines and
into the low \(20\%\) range for masked runs.  Reasoning models still have higher pass rates.  

\textit{Take-away:} Concise, semantically pivotal edits force actual
rewrites; the anchor cannot override an explicit change in task
polarity in most cases.

\paragraph{Across-model heterogeneity.}
Overall, model by model trend analysis show varying responses and pass rate dynamics.  Such idiosyncrasies likely trace back to different training mixtures and
alignment objectives.  Stepwise curves therefore provide a richer
picture of model quality than any single aggregate score.

Overall, the disaggregated trajectories reveal three robustness regimes:

\begin{itemize}[leftmargin=1.4em,nosep]
  \item \textbf{Text-robust but anchor-dependent} (PD–Unmasked);
  \item \textbf{Anchor-sensitive but text-stable} (early JI);
  \item \textbf{Anchor-agnostic but semantics-fragile} (LF).
\end{itemize}

These complementary failure modes highlight why selective robustness—
\emph{tolerate benign noise, react to critical edits}—remains an open
challenge for even the most advanced LLMs.

\subsection{External and internal model consensus}\label{sec:consensus}

\begin{figure*}[t]
    \centering
    \includegraphics[width=.95\textwidth]
        {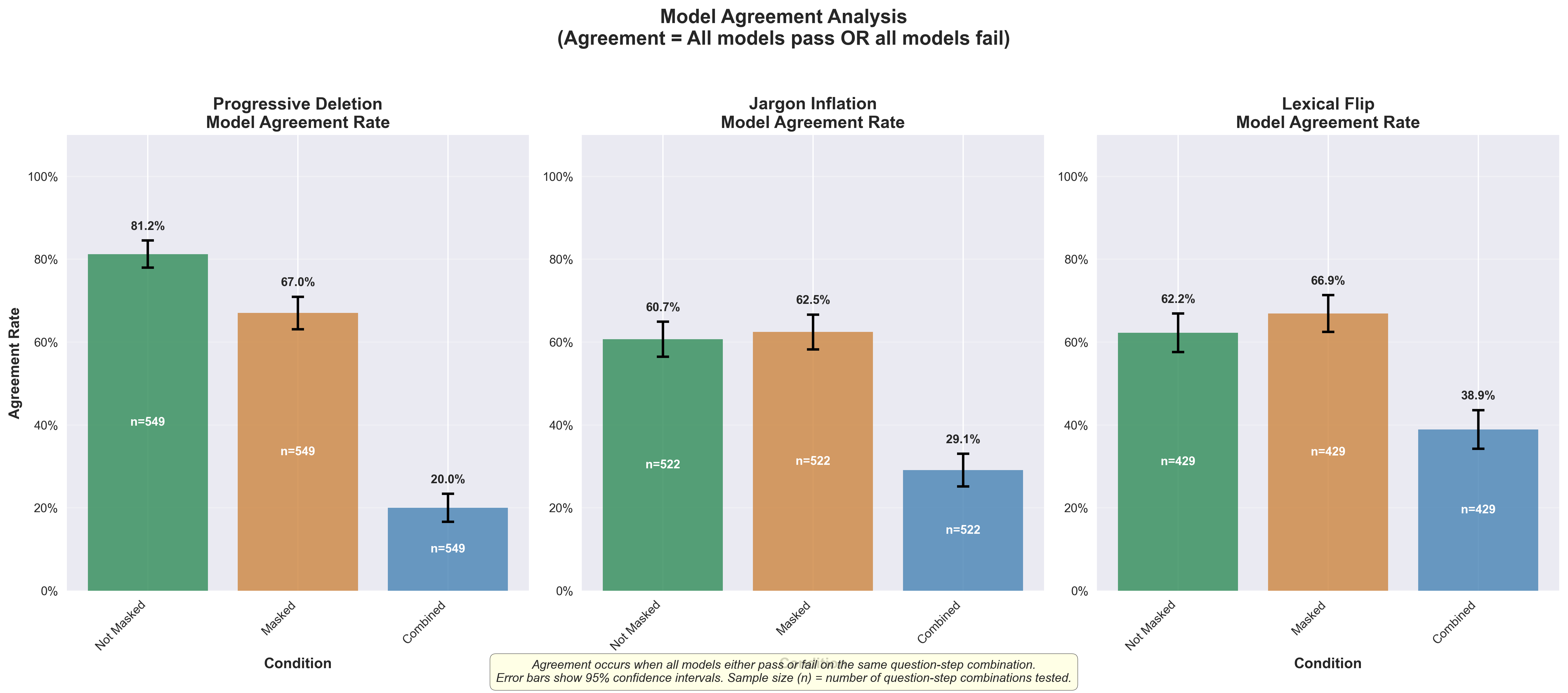}
    \caption{Strong agreement between models.
         For each perturbation family and identifier mode, the bar height
         is the proportion of question–step pairs in which \emph{all six
         models gave the same verdict} (either all pass or all fail).     
         The numeral printed above each bar is that percentage; the
         italic $n$ below gives the number of pairs in the group.
         Error bars show bootstrapped 95\,\% confidence intervals.}
    \label{fig:agreement-overview}
\end{figure*}

Figure \ref{fig:agreement-overview} shows the strong-agreement rate for every perturbation family (PD, JI, LF) and identifier mode (Unmasked, Masked, Combined).
For each question at each perturbation step we look at the all models’ test-suite outcomes (pass or fail). If all models give the same verdict—either every one passes or every one fails—we call that a strong-agreement case. We tally the number of strong-agreement cases and divide by the total number of question–step pairs in that group, giving a percentage.The vertical lines represent 95 \% confidence intervals obtained by bootstrapping the question–step pairs within each bar.
\smallskip

\textbf{Combined case.}  
Each problem–step is evaluated twice by the \emph{non-reasoning}
variants (Masked \& Unmasked) and once by the \emph{reasoning} variant
(Masked only).  A strong-agreement count of~1 therefore implies that the
non-reasoning model reaches the \emph{same} verdict with and without the
anchor, and that the reasoning models concur.

\paragraph{Headline pattern.}
When we keep \emph{one} identifier setting (Masked \emph{or} Unmasked)
the six models often agree: strong‐agreement rates cluster between
\textbf{60}\,–\,\textbf{81 \%}.  
The picture changes once we pool the two settings
(\emph{Combined}): agreement collapses to \textbf{20}\,–\,\textbf{39 \%}.
Pooling forces every non-reasoning model to solve the \emph{same} prompt
twice—first with its familiar anchor, then without—and strong agreement
now requires it to make the \emph{same} decision both times \emph{and}
match the reasoning variant.  The sharp drop therefore exposes
intra-model anchor dependence as well as inter-model divergence.

\paragraph{Family-level detail.}
\begin{enumerate}[label=\arabic*.,leftmargin=1.6em,nosep]
  \item \textbf{Progressive Deletion (PD).}  
        Highest unanimity in the unmasked run (\textbf{81 \%}); masking
        removes the compass and consensus slips to 67 \%.  Mixing the
        two modes drives it down to \textbf{20 \%}.
  \item \textbf{Jargon Inflation (JI).}  
        Anchor presence matters little (61 \% vs.\ 63 \%), confirming
        that most models ignore exotic terms.  The combined rate still
        halves to 29 \%, driven mainly by anchor conflict inside each
        baseline model.
  \item \textbf{Lexical Flip (LF).}  
        Lowest cohesion when the anchor is present (62 \%) because the
        subtle polarity swap splits models.  Masking nudges consensus up
        to 67 \%, yet pooling both modes cuts it back to 39 \%.
\end{enumerate}

\paragraph{Interpretation.}
Two opposing forces govern consensus:

\begin{itemize}[leftmargin=1.4em,nosep]
  \item \textbf{Anchor alignment.}  
        A shared descriptive name steers different models toward the
        same canonical template, inflating agreement.
  \item \textbf{Anchor conflict.}  
        Presenting each model with both an anchored and an anchor-free
        version of the \emph{same} problem reveals hidden heuristics and
        slashes agreement.
\end{itemize}

\smallskip
Instruction following is therefore \textbf{highly non-uniform}.  A
single identifier can swing strong-agreement rates by more than
30 pp.  Robust evaluation must audit both anchored and anchor-free
prompts and report ensemble-level consistency, not single-model
snapshots.

\subsection{Semantic drift in chain-of-thought}\label{sec:sem-drift}

\begin{figure*}[t]
    \centering
    \includegraphics[width=.9\textwidth]
        {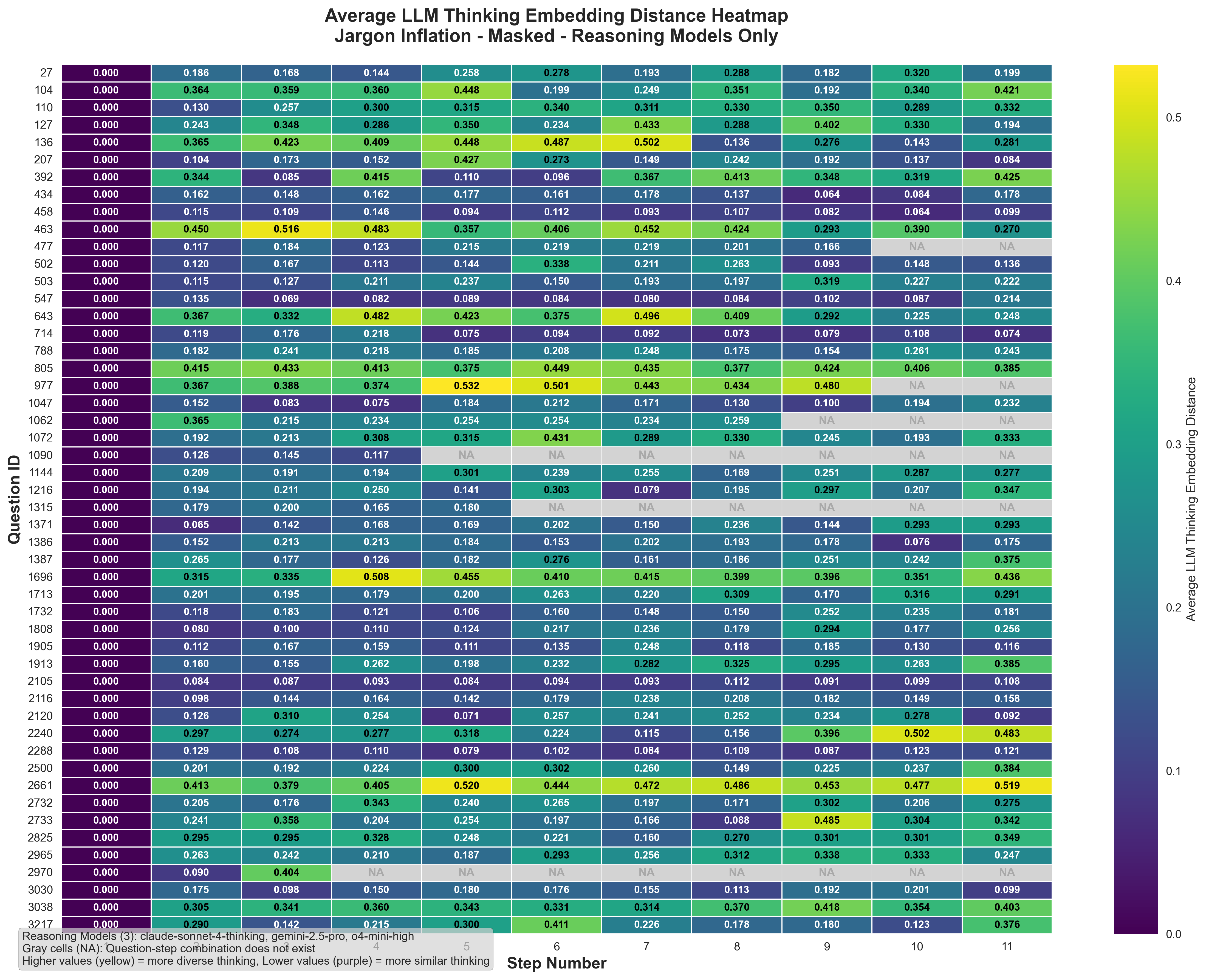}
    \caption{Chain-of-thought drift for the reasoning variants in the
             \textbf{Jargon-Inflation + Masked} track.
             Darker cells = greater textual change; grey = invalid
             generation.}
    \label{fig:thinking-heatmap}
\end{figure*}

For each prompt we embed the natural-language rationale produced at
\textit{every} perturbation step and measure its cosine distance to the
step-0 rationale (Figure \ref{fig:thinking-heatmap}).  
Embeddings come from Sentence-BERT; all observed distances fall in the
\([0,1]\) range, where 0 means “verbatim” and values above about 0.30
sound noticeably different to a human reader.

\paragraph{Key observations.}
\begin{enumerate}[label=\arabic*.,leftmargin=1.6em,nosep]
  \item \textbf{Anchor removal is the biggest trigger.}  
        The very first step—masking the function name—pushes the median
        distance from 0 to roughly 0.24, before any jargon appears.
  \item \textbf{Jargon adds further drift.}  
        Distances creep upward with each extra jargon substitution,
        reaching about 0.35 for the upper quartile by step 10.  The
        prose changes a lot even when the code underneath does not.
  \item \textbf{New words not equal to new behaviour.}  
        More than half of the rationales that drift by 0.30 or more
        still pass the original test suite.  The model is talking as if
        the task changed while quietly re-using the old algorithm.
\end{enumerate}

\paragraph{Implications.}
Reasoning-optimised endpoints are skilled at rephrasing their
explanations once anchor cues disappear, but this rhetorical flexibility
rarely coincides with genuine algorithmic adaptation.  Large semantic
drift in chain-of-thought should therefore be treated as \emph{narrative
noise}, not as evidence that the model has truly understood and
implemented the new requirement.

\subsection{Factor analysis}\label{sec:factor}

\paragraph{Objective.}
Which prompt edits make canonical regression \emph{more} likely?
For every question–step pair we compute nine surface- and semantic-level
features (Table~\ref{tab:features}).  
We then contrast their distributions between \textit{pass} (regression)
and \textit{fail} groups using Cohen's~$d$.\footnote{Positive
$d$\,\(\rightarrow\) the feature is larger when the model
\emph{adapts} to a new task; negative when it \emph{fails} to adapt.}

\begin{figure*}[t]
    \centering
    \includegraphics[width=\linewidth]
        {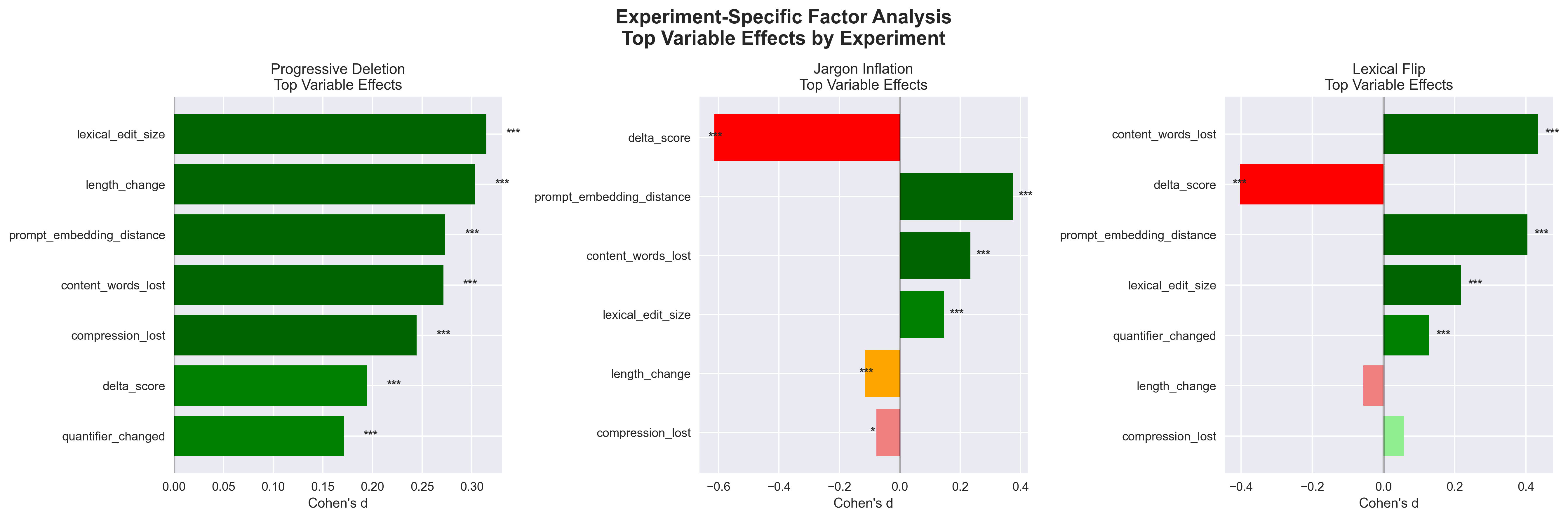}
    \caption{Top predictors of pass/fail by perturbation family.
             Green bars: feature promotes regression ($d>0$); red bars:
             feature promotes failure ($d<0$).
             Significance: {\small\textbf{***}\,$p{<}.001$,
                         \textbf{**}\,$p{<}.01$,
                         \textbf{*}\,$p{<}.05$}.}
    \label{fig:factor-panels}
\end{figure*}

\paragraph{Composite semantic shift $(\Delta)$.}
To capture global prompt drift we define
\[
\begin{aligned}
\Delta \;=\;& 0.40\,\text{CLS--EMD}\\
           &+\,0.60\!\bigl(1-\text{BERTScore}_{F1}\bigr)\\
           &+\,0.40\,\text{NLI}_{\text{contradict}}\\
           &+\,0.70\,\text{span\_}\Delta .
\end{aligned}
\]

where
\begin{itemize}[leftmargin=1.6em,nosep]
  \item \textbf{CLS--EMD}: Earth–Mover’s distance between sentence-level
        \verb|[CLS]| embeddings;
  \item \textbf{BERTScore$_{F1}$}: lexical overlap (higher $\Rightarrow$
        closer);
  \item \textbf{NLI$_{\text{contradict}}$}: contradiction/neutrality
        probability from an MNLI-tuned RoBERTa;
  \item \textbf{span\_}$\Delta$: mean cosine gap across equal-length
        sentence spans.
\end{itemize}

\paragraph{Key findings (Fig.\,\ref{fig:factor-panels}).}
\begin{enumerate}[label=\arabic*.,leftmargin=1.6em,nosep]
  \item \textbf{Progressive Deletion.}  
        Token-loss metrics
        (\textit{lexical\_edit\_size}, \textit{length\_change},
        \textit{content\_words\_lost}) dominate ($d\approx0.30$,
        $p<.001$).  The more text disappears, the less safer the model feels
        reusing its template.
  \item \textbf{Jargon Inflation.}  
        The $\Delta$-score is the strongest predictor ($d=-0.61$). Large
        esoteric shifts \emph{increase} pass-rates or fails to adapt: faced with obscure
        jargon, models choose to ignore it.
  \item \textbf{Lexical Flip.}  
        Here $\Delta$ turns \emph{negative} ($d=-0.40$) while features
        tied to the specific edit
        (\textit{content\_words\_lost}, \textit{prompt\_embedding\_distance})
        are positive.  A tiny but pivotal word change is noticed—and
        often mishandled.
\end{enumerate}

\paragraph{Practical takeaway.}
LLMs fail not on verbose complexity but on \emph{concise,
meaning-critical edits}.  Future stress tests should therefore focus on
minimal, semantics-altering perturbations (e.g.\ quantifier or unit
flips) rather than wholesale rewrites.

\begin{table*}[t]
\centering
\small
\caption{Feature definitions used in the factor analysis.  Metrics above
the horizontal rule depend \emph{only} on the prompt/edit pair; the last
two rely on the generated code or embeddings.}
\label{tab:features}
\begin{tabular}{p{4.4cm}p{11cm}}
\toprule
\textbf{Feature} & \textbf{Description and range} \\
\midrule
\textit{length\_change} & Fractional change in prompt length
  (tokens). $+1$ = all tokens removed, $0$ = no change, $-1$ = length doubled. \\[2pt]

\textit{lexical\_edit\_size} & Normalised token-level Levenshtein distance
  ($0\!-\!1$). \\[2pt]

\textit{compression\_loss} & Difference in gzip compression ratio between
  \textit{edit} and \textit{base} ($-1\!-\!1$).  
  Positive $\Rightarrow$ the edited prompt is \emph{less} compressible. \\[2pt]

\textit{content\_words\_lost} & Share of content tokens (POS $\in$
  \{N, V, ADJ, ADV\}) present in the base prompt but absent in the edit
  ($0\!-\!1$). \\[2pt]

\textit{negation\_deleted} & Boolean: any negator
  (\emph{no, not, never, n't}) appears in the base but not in the edit. \\[2pt]

\textit{negation\_added} & Boolean: any negator appears in the
  edit but not in the base. \\[2pt]

\textit{quantifier\_changed} & Boolean: at least one quantifier
  (\emph{all, any, most, none, some, few, many, one, two, …}) is added or
  removed. \\
\midrule
\textit{prompt\_embedding\_distance} & SBERT cosine distance between base
  and edited prompt ($0\!-\!2$; here $<1$). Higher = greater semantic drift. \\[2pt]

\textit{delta\_score} & Composite Semantic Distance Index
  ($0$ identical, $1$ disjoint). \\
\bottomrule
\end{tabular}
\end{table*}

\section{Limitations}\label{sec:limits}

The study reveals consistent patterns, yet several factors restrict how far the conclusions can be generalised:

\begin{itemize}[leftmargin=1.5em,nosep]
    \item \textbf{Task coverage.}  
          We analyse a sample of \textit{50} coding problems—less than 2\,\% of the full \textsc{LeetCodeDataset}.  
          A larger sample, and domain-diverse corpus (e.g.\ maths proofs, chain-of-thought riddles, multimodal tasks) might expose different failure modes and improve generalizability of the findings.

    \item \textbf{Evaluation oracle.}  
          Solutions are graded against the \emph{unaltered} LeetCode test suites.  
          A passing score therefore demonstrates \emph{compatibility with the original task}, not correctness for the mutated one. Moreover, new questions might have been compatible with those tests. Constructing fresh ground-truth outputs for every perturbation would tighten causal attributions.
    \item \textbf{Code-only enforcement.}  
          We forced all models to return Python code. Under severe underspecification an LLM might naturally ask clarifying questions or refuse; we suppress that behaviour, potentially underestimating pragmatic robustness.  

    \item \textbf{Original Leetcode problem traces and structure}  
          While we could measure the effect of changing function name, other traces of the problem remained, such as class name ``Solution'' or function signatures that may strongly predict the original LeetCode problem. Cleanup might have been not deep enough to clean within the problem description examples, therefore creating constraint contradictions. We decided to keep the problems as they are, as those highlight the question we are raising here: what factors keep LLMs returning to the canonical problem? 

    \item \textbf{Reasoning variant scope.}  
          “Reasoning’’ endpoints were sampled only in the \textit{masked} tracks only to control research costs.  Evaluating them across all six tracks—and with multiple temperature seeds—would clarify whether the small accuracy lift we observe is systematic.
    
    \item \textbf{True solutions}  
          Probably under different query prompt frontier LLMs could have solved all updated problems, but that was out of the scope of this research as we tried to test sensitivity to marginal symbolic changes made to canonical problem. 

\end{itemize}

\section{Literature Review}\label{sec:related}

The tendency of large language models (LLMs) to fall back on
high-probability training patterns when prompts are noisy or
underspecified echoes earlier observations of over-generalisation in
sequence-to-sequence systems.  
In abstractive summarisation, even minor edits to the source can trigger
factual hallucinations~\citep{Maynez2020}.  
Likewise, inserting a single distractor sentence into a reading
comprehension passage sharply lowers QA accuracy, revealing reliance on
spurious cues~\citep{JiaLiang2017}.  
These studies imply that token-level likelihood often outweighs semantic
fidelity.

\textbf{Prompt sensitivity} sharpens the lens.  
Surface paraphrases alone can make GPT-3.5 oscillate between
contradictory answers~\citep{Ye2023Robust}.  
Chain-of-thought prompting boosts multi-step reasoning but only while
key tokens remain visible~\citep{Wei2022CoT}.  
Adversarial triggers show the extreme case: nonsense token sequences can
hijack model output, indicating reliance on superficial
patterns~\citep{Wallace2019Triggers}.  
Our anchor-masking results parallel this: removing a single function
name slashes pass-rates, exposing the shortcut.

\textbf{Evaluation frameworks} now systematise such stress tests.
BIG-Bench probes generalisation across 200+ tasks under distribution
shift~\citep{Srivastava2022BigBench}, while HELM offers a robustness
dashboard spanning performance, fairness, and calibration under prompt
variation~\citep{Liang2022HELM}.  
Ensemble diagnostics further uncover hidden bias: combining a “biased”
and a standard model reveals artefact reliance that single systems hide
\citep{Clark2019Dont}.  
Our perturbation suite extends this agenda to algorithmic code
generation, contrasting benign noise (underspecification), ambiguous
drift (jargon), and meaning-critical edits (quantifier flips) to gauge
whether models exhibit \emph{good} or \emph{bad} sensitivity.

\section*{Conclusion}

\paragraph{Main insights.}
Across six frontier models, three perturbation families, and
11\,853 generations, we observe three consistent patterns:

\begin{enumerate}[label=\arabic*.,leftmargin=1.6em,nosep]
  \item \textbf{Identifier anchoring.}  
        A single descriptive function name dominates task inference:
        masking it cuts pass-rates by up to \textbf{20 pp}, irrespective
        of how much of the prompt remains.

  \item \textbf{Regression under drift.}  
        When 90 \% of tokens are deleted \emph{or} jargon is injected,
        roughly \textbf{85 \%} and \textbf{56 \%} of generations respectively still pass the \emph{original}
        tests—evidence that models revert to memorised templates rather
        than parsing noisy detail.

  \item \textbf{Tiny edits, huge impact.}  
        One polarity or unit swap can flip task semantics, yet fewer than
        half of the models adapt.  These “minimal but maximal” edits
        expose the brittle edge of current instruction following.
\end{enumerate}

Reasoning-optimised endpoints do not solve the problem; they add longer
explanations to the same canonical code.  
Once anchored and anchor-free prompts are mixed, strong agreement across
(and within) models drops below 40 
prompt’’ is far from a closed issue.

\paragraph{Implications.}
Robust deployment requires breaking anchor reliance and rewarding
token-level fidelity.  Benchmarks that keep the anchor intact yet vary
other surface details \emph{overstate} current capability; sensitivity
tests must include concise, meaning-critical edits.

\paragraph{Next steps.}
\begin{itemize}[leftmargin=1.4em,nosep]
  \item \textbf{Anchor randomisation} in both training and evaluation to
        prevent brittle heuristics.
  \item \textbf{Minimal-edit adversaries} (unit swaps, quantifier flips)
        as a permanent red-team fixture.
  \item \textbf{Perturbation-aware oracles}: grade models on the
        \emph{mutated} task, not the canonical one.
  \item \textbf{Architectural work}: retrieval-conditioned decoding or
        token-aligned rewards that force re-parsing instead of template
        recall.
\end{itemize}

Moving from surface plausibility to fine-grained comprehension is
essential before LLMs can be trusted in high-stakes, rapidly changing
environments.

\section*{Paper's Dataset}

 The dataset is available at \url{https://huggingface.com/level8/leetp}.

\bibliographystyle{plainnat}

\section*{Appendix}

\subsection*{Appendix A – General LLM prompt template}

\lstset{basicstyle=\ttfamily\scriptsize,breaklines=true}
\begin{lstlisting}[language=Python]
You are an expert Python programmer. You will be given a question and will generate a correct Python program.

### Question:
{cleaned_problem_description}

### Format: Write the solution to the problem and enclose your code within delimiters.

```python
{the_starter_code}
Answer: (use the provided format with backticks)

\end{lstlisting}

\subsection*{Appendix B – “Jargon Inflation’’ meta-prompt}

\begin{lstlisting}[language=Python]
You are a top Computer Science Professor who is creating a tricky exam question for students.
Since you gave away all previous exam questions to students to prepare, you want to update the last exam question such that it looks very similar, may have similar solution, but will not pass the same test cases. You want to test students` understanding, not memorization.

The last exam question: {original_paragraph}

The previous exam questions:
{history_str}

Substitute EXACTLY ONE WORD such that it changes the last exam question in a subtle and non-obvious way while not repeating previous exam questions. The change MUST reflect
HIGHLY advanced TECHNICAL and OBSCURE concepts!

Return the updated exam question ONLY, without hints or explanation.
\end{lstlisting}

\subsection*{Appendix C – “Lexical Flip’’ meta-prompt}

\begin{lstlisting}[language=Python]
You are a top Computer Science Professor who is creating a tricky exam question for students.
Since you gave away all previous exam questions to students to prepare, you want to update the last exam question such that it looks very similar, may have similar solution, but will not pass the same test cases. You want to test students` understanding, not memorization.

The last exam question: {original_paragraph}

The previous exam questions:
{history_str}

Substitute EXACTLY ONE WORD such that it changes the last exam question in a subtle and non-obvious way while not repeating previous exam questions.

Return the updated exam question ONLY, without hints or explanation.
\end{lstlisting}

\subsection*{Appendix D – Figures and discussion}

\subsubsection*{Difficulty levels}

\begin{figure*}[t]
\centering
\includegraphics[width=.75\textwidth]
{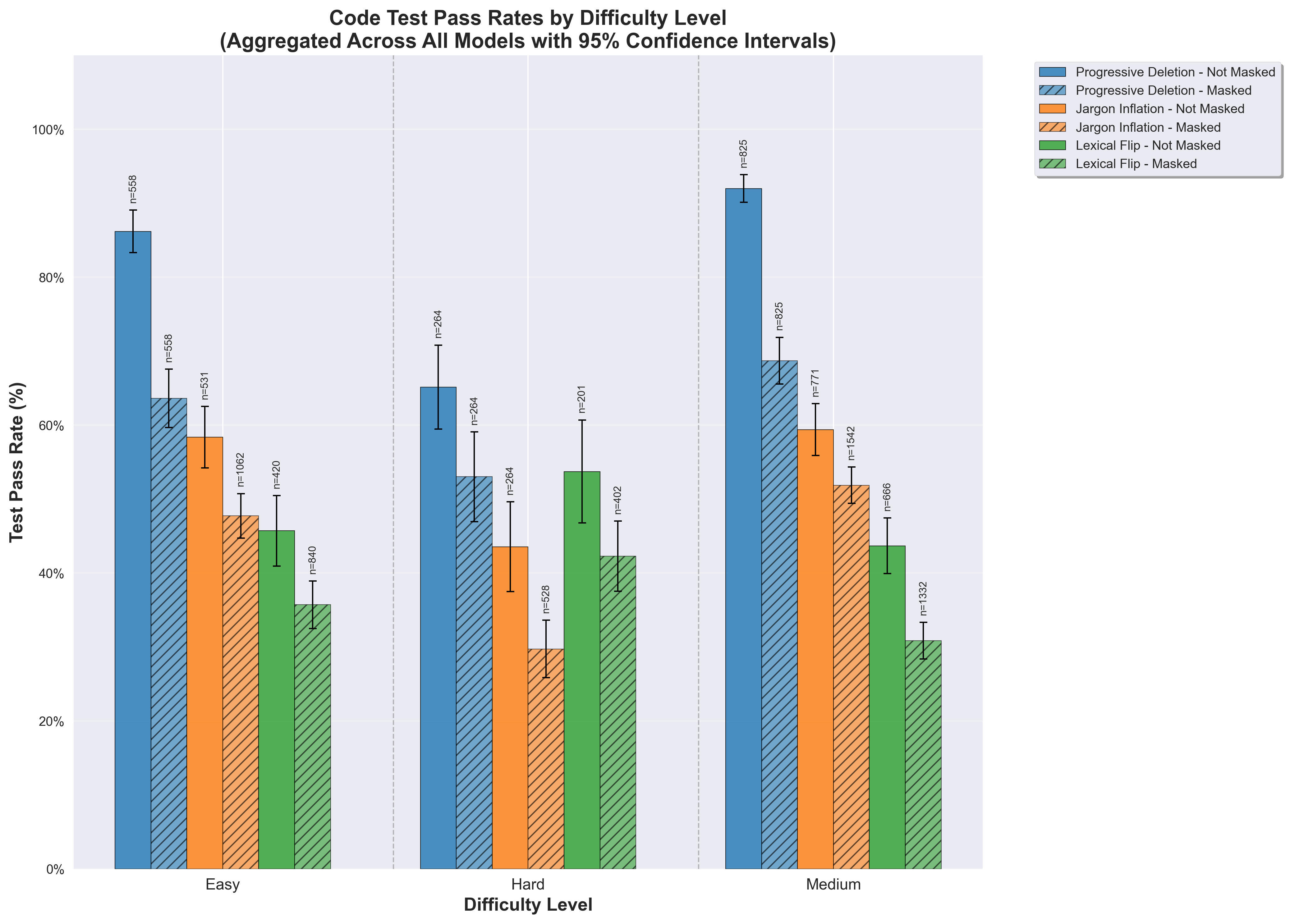}
\caption{Pass-rate by LeetCode difficulty tier.}
\label{fig:difficulty}
\end{figure*}

Figure \ref{fig:difficulty} disaggregates pass-rates by LeetCode
difficulty tier.  \textbf{Easy} and \textbf{Medium} problems remain
uniformly high across tracks.  In the \textbf{Hard} tier, however,
\textbf{Progressive deletion} falls sharply, \textbf{Lexical Flip}
retains a modest edge, and \textbf{Jargon Inflation} triggers the
largest—and significant—drop.

\subsubsection*{Halstead difference}

\begin{figure*}[t]
\centering
\includegraphics[width=.75\textwidth]
{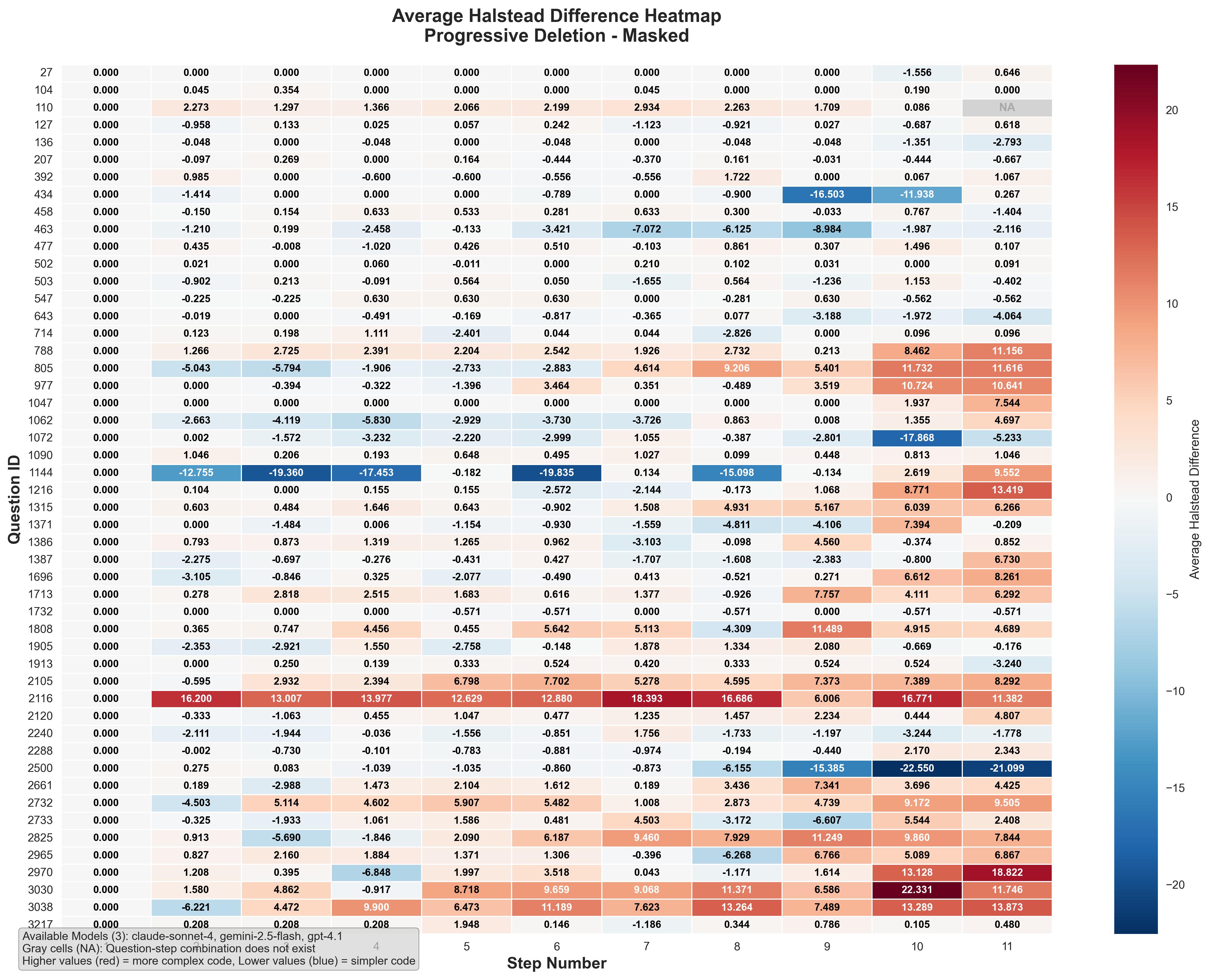}
\caption{Average change in Halstead Difficulty (baseline – step).}
\label{fig:halstead}
\end{figure*}

We quantify symbolic change with Halstead complexity
\citep{Halstead1977}.  Figure \ref{fig:halstead} contrasts the baseline
code with successive perturbation steps on the
\textbf{Jargon Inflation – Masked} track.  Halstead Difficulty rises
monotonically, showing progressive simplification:

\[
\begin{aligned}
\Delta\text{Difficulty}
  &= \text{Difficulty}_{\text{base}}
     \;-\;
     \text{Difficulty}_{\text{step}} .
\end{aligned}
\]

Positive values dominate in later steps, confirming systematic
simplification.  Unmasking the identifier amplifies this effect across
all models.

\subsubsection*{Model Comparison}

\begin{figure*}[t]
\centering
\includegraphics[width=.75\textwidth]
{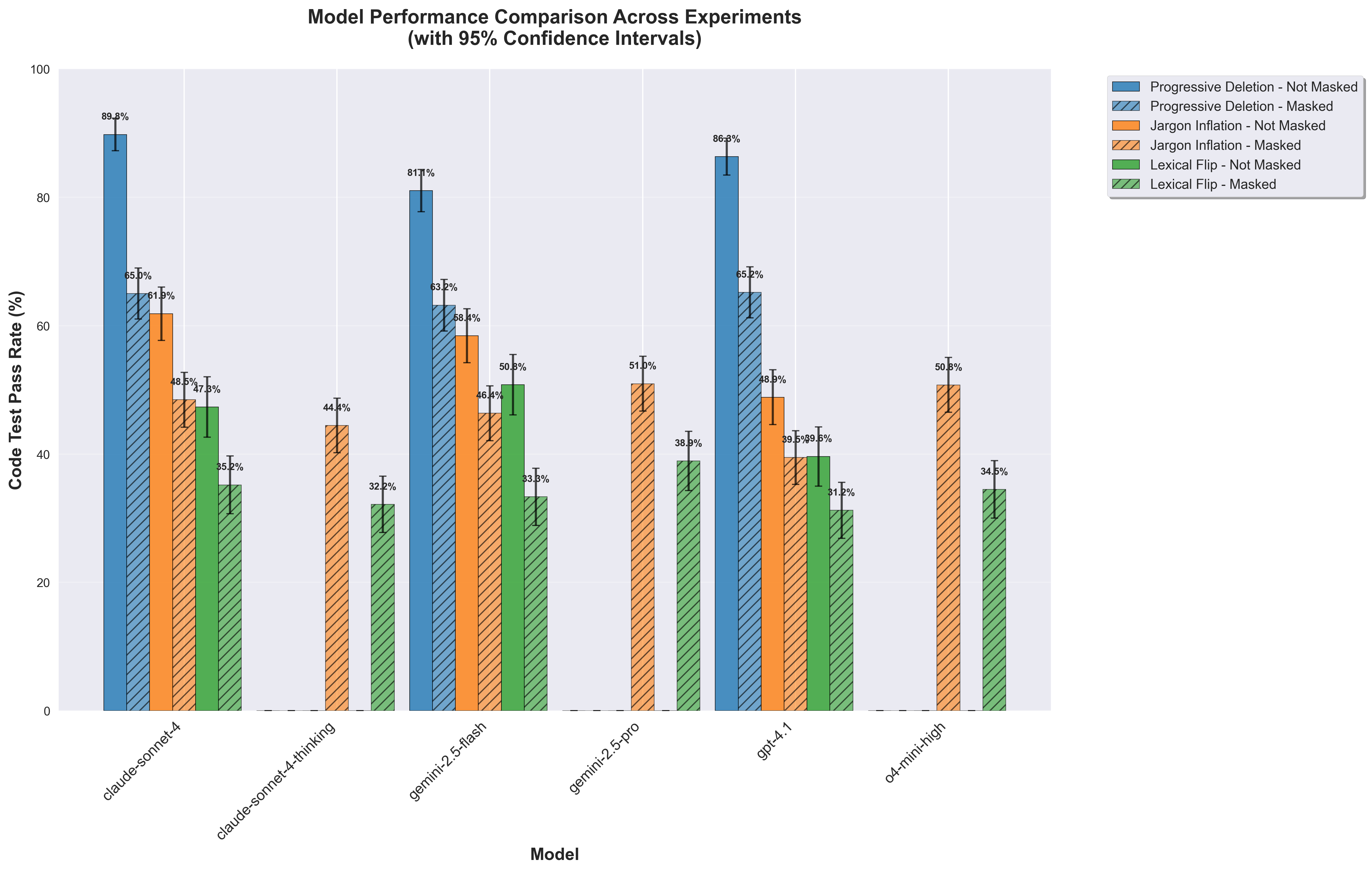}
\caption{Model performance comparison accross experiments.}
\label{fig:comparison}
\end{figure*}

In Figure \ref{fig:comparison} we see consistency in model performance across experiments, removing anchors resulting in significant drops in pass rates.  

\end{document}